\RequirePackage{fix-cm}

\documentclass[twocolumn]{svjour3}          

\smartqed  

\usepackage{color,graphicx}
\usepackage{multirow,multicol}
\usepackage[utf8]{inputenc}
\usepackage{amssymb}
\usepackage{amsmath}
\usepackage{subfigure}
\usepackage{url}
\usepackage{xfrac}

\newcommand{\sign}{\text{sign}}

\journalname{Neural Computing \& Applications}

\begin{document}
\onecolumn
\title{Enhanced perceptrons using contrastive biclusters}

\author{André L. V. Coelho \and Fabrício~O.~de~França}

\institute{André L. V. Coelho \at
              Graduate Program in Applied Informatics, Center of Technological Sciences, University of Fortaleza (UNIFOR), Brazil \\
              \email{acoelho@unifor.br}           
          \and
           Fabrício O. de França \at
              Center of Mathematics, Computing and Cognition (CMCC), Federal University of ABC (UFABC), Brazil \\
			\email{folivetti@ufabc.edu.br}
}

\date{Received: date / Accepted: date}
\twocolumn
\maketitle

\begin{abstract}
Perceptrons are neuronal devices capable of fully discriminating linearly separable classes. Although straightforward to implement and train, their applicability is usually hindered by non-trivial requirements imposed by real-world classification problems. Therefore, several approaches, such as kernel perceptrons, have been conceived to counteract such difficulties. In this paper, we investigate an enhanced perceptron model based on the notion of {\em contrastive biclusters}. From this perspective, a good discriminative bicluster comprises a subset of data instances belonging to one class that show high coherence across a subset of features and high differentiation from nearest instances of the other class under the same features (referred to as its contrastive bicluster). Upon each local subspace associated with a pair of contrastive biclusters a perceptron is trained and the model with highest area under the receiver operating characteristic curve (AUC) value is selected as the final classifier. Experiments conducted on a range of data sets, including those related to a difficult biosignal classification problem, show that the proposed variant can be indeed very useful, prevailing in most of the cases upon standard and kernel perceptrons in terms of accuracy and AUC measures.
\keywords{Perceptrons \and Discriminative biclustering \and Contrastive patterns \and Kernels \and Binary classification~\footnote{This article is under review by Neural Computing and Applications, Springer}}
\end{abstract}

\section{Introduction}
\label{sec:intro}

Given a data matrix $\mathbf{X}=[R,C]$ composed of $N=|R|$ rows (objects/instances) and $M=|C|$ columns (features/attributes), a  {\em bicluster} (or co-cluster) $\mathbf{B}=[I,J]$ can be defined as a submatrix of $\mathbf{X}$ whose elements $\mathbf{B}(i,j), i \in I \subseteq R, j \in J \subseteq C,$ show a high level of {\em similarity} (homogeneity) among themselves. Such definition generalizes the well-known concept of a cluster~\cite{jain2010} in the sense that now instances can be grouped together considering only subsets of the features, and vice-versa, allowing for the elicitation of more contextualized data models~\cite{madeira2004}. Since different notions of similarity and bicluster representation can be adopted, different bicluster models have been investigated in the preceding years, giving birth to a number of biclustering algorithms deployed in several application domains~\cite{dhillon2001,dhillon2003,han2009,zhao2012,huang2011,huang2015,defranca2013,defranca2015}.

Besides the bicluster model, biclustering algorithms usually differ on the strategy adopted to recover the final set of biclusters. One common solution is alternating between the clustering of rows and columns of the data set, allowing the use of traditional clustering algorithms. Another strategy is to employ greedy heuristics that iteratively insert or remove rows/columns into the bicluster so as to maximize a given objective function. Also noteworthy are some probabilistic approaches seeking for a generative model underlying the biclusters. Divide-and-conquer methods are especially useful to finding a block structure inside the data set. On the other hand, some population-based meta-heuristics have also been adopted for optimizing a criterion function that promotes a balance between the size of the bicluster and the homogeneity of its elements. Finally, when small or medium sized data sets are in sight and some constraints, such as no overlapping, are satisfied, enumerative approaches may be used to extract the full set of biclusters. For a comprehensive overview of these approaches, we refer the reader to~\cite{madeira2004,freitas2013}.

A well-known class of biclusters is that based on the notion of {\em coherence}, whereby the values of $\mathbf{B}$ correlate to each other according to an implicit linear (additive or multiplicative) model~\cite{cheng2000,odibat2014}. One asset of this particular bicluster type is that it can simultaneously capture different coassociation patterns among the instances and features of a matrix, generalizing other simpler bicluster types. Another advantage is that the level of coherence revealed by a given bicluster can be efficiently measured via a simple similarity function, known as the {\em mean-squared residue} (MSR) score~\cite{cheng2000,huang2011}.

Conventional bicluster analyses are unsupervised in nature, meaning that the class labels possibly associated with the instances (or features) are not taken into account. More recently, another class of algorithms, referred to as {\em discriminative} (or differential, supervised) biclustering algorithms, has gained increased  attention \cite{odibat2014,fang2010,wang2013,defranca2015}.  These methods somehow incorporate class label details into the process of eliciting biclusters so as to produce better predictive models. Assuming, for instance, the context of gene expression data analysis, discriminative biclustering has been used to discover sets of genes (instances) that are correlated (co-expressed) under a subset of experimental conditions (features) pertaining to one class of conditions (e.g., normal tissues) but not to the other (e.g., cancerous tissues)~\cite{odibat2010}. Usually, discriminative biclustering methods differ from each other according to the different strategies they apply for identifying differential patterns among the considered classes.

In this paper, we propose a novel supervised biclustering approach, named as BicNeuron, aiming at detecting highly discriminative coherent biclusters for leveraging the accuracy and generalization of simple linear classification models. Our focus will be specifically on perceptrons, one of the most studied classes of neuronal devices for which several variants have been proposed. Among them, kernel perceptrons~\cite{freund1999} have achieved notoriety for their elegant strategy in dealing with non-linearities via kernel functions~\cite{scholkopf2002}. As far as we are aware of, this is the first study investigating the usefulness of combining discriminative biclusters with simple neural models.

In a nutshell, BicNeuron seeks out for subsets of training instances belonging to one of the classes of instances, showing not only high coherence across subsets of features but also high differentiation from the nearest instances of the opposite class under the same features. This procedure may yield several local discriminative subspaces captured as pairs of {\em contrastive biclusters}. Upon data of each local subspace, a (standard or kernel) perceptron model is induced and the one with highest value of the area under the receiver operating characteristic curve (AUC) is selected as the final model for classifying novel test instances.

In what follows, we outline in Section~\ref{sec:back} the theoretical background behind BicNeuron, whose steps are detailed in Section~\ref{sec:bicneuron}. Next, we discuss the results achieved in experiments conducted on a range of data sets, also including a case study on a difficult biomedical classification problem~\cite{lima2011,nunes2014}. Finally, Section~\ref{sec:conclusion} concludes the paper and brings remarks on future work.

\section{Background and related work}
\label{sec:back}
In the sequel, we briefly revise the main training steps behind standard and kernel perceptrons. Then, we provide an account of some concepts and approaches related to coherent and discriminative biclustering.

\subsection{Standard and kernel perceptrons}
\label{sec:perc}

Let $\mathbf{X}$ be a set of $N$ training instances $\mathbf{x}_1,\ldots,\mathbf{x}_N$ and their corresponding labels $y_1,\ldots,y_N$, where $y_k \in \{-1,$ $+1\}$. The standard perceptron model~\cite{alpaydin2014} predicts the class $y_k$ of a certain instance $\mathbf{x}_k$ with the function
\begin{equation}
f(\mathbf{x}_k) = \sign(\mathbf{w}\cdot\mathbf{x}_k),
\label{eq:perc1}
\end{equation}
where $\mathbf{w}$ is the set of free parameters (including a bias term) defining the decision boundary and $\sign(\mathbf{w}\cdot\mathbf{x}_k)$ is +1, if $\mathbf{w}\cdot\mathbf{x}_k>0$, and -1, otherwise. When the class prediction of $\mathbf{x}_k$ is correct, no adjustment to $\mathbf{w}$ is required; otherwise, $\mathbf{x}_k$ is added to (subtracted from) $\mathbf{w}$ if $y_k = +1 (-1)$. That is, the update rule from iteration $t$ to iteration $t+1$ can be simply written as
\begin{equation}
\mathbf{w}(t+1) = \mathbf{w}(t)+\beta y_k\mathbf{x}_k,
\label{eq:perc2}
\end{equation}
where $\beta$ denotes the learning rate, whose role is to modulate the effect of the adjustment term on the new values of the weights.

From \eqref{eq:perc2}, one can notice that when $\beta=1$ and $\mathbf{w}$ is initialized to the null vector, the latter will then become a simple linear combination of the training instances, that is, $\mathbf{w}=\sum_l \alpha_l y_l \mathbf{x}_l, l= 1,\ldots,N$. Reconsidering \eqref{eq:perc1} from this perspective and assuming that $\phi(\mathbf{x}_k)$ denotes some linear/non-linear transformation applied to $\mathbf{x}_k$, one can derive a kernel-based type of perceptron~\cite{freund1999}:
\begin{eqnarray}
f(\mathbf{x}) &=& \sign(\sum_l \alpha_l y_l \phi(\mathbf{x}_l)\cdot\phi(\mathbf{x}_k)) \nonumber\\
&=& \sign(\sum_l \alpha_l y_l K(\mathbf{x}_l,\mathbf{x}_k)),
\end{eqnarray}
where $K(\mathbf{x}_l,\mathbf{x}_k) = \phi(\mathbf{x}_l)\cdot\phi(\mathbf{x}_k)$ denotes a particular type of similarity function referred to as a Mercer kernel~\cite{scholkopf2002}. For kernel perceptrons, the update rule for when a mistake is made predicting the class of $\mathbf{x}_k$ becomes 
\begin{equation}
\alpha_k(t+1) = \alpha_k(t) + 1,
\end{equation}
with $\alpha_k$ denoting the number of times $\mathbf{x}_k$  was mistaken (that is, how difficult it is to correctly classify this instance).

\subsection{Coherent and discriminative biclustering}
\label{sec:bics}

An unsupervised approach that may be useful for finding subspaces of linearly separable data is the additive coherence biclustering~\cite{cheng2000}. According to this approach, a subset of the data instances show a strong linear coassociation pattern when viewed through the perspective of a subset of their features. In other words, each instance vector of a bicluster presents a similar profile, except for a constant bias, when projected onto the selected coordinates of the bicluster. In this case, each element $\mathbf{B}(i,j)$ of the submatrix associated with the bicluster can be modeled as:
\begin{equation}
\mathbf{B}(i,j) = \mathbf{B}(I,j) + \mathbf{B}(i,J) - \mathbf{B}(I,J),
\end{equation}
where $\mathbf{B}(I,j)$ is the mean value of the $j$-th column of $\mathbf{B}$, $\mathbf{B}(i,J)$ is the mean value of the $i$-th row, and $\mathbf{B}(I,J)$ is the mean value of the whole bicluster. 

The levels of coherence in a bicluster can be measured by the {\em MSR score}, which, for additive coherence, can be written as:
\begin{equation}
H(I,J) = \frac{1}{|I||J|}\sum_{i,j \in I,J}{ r(i,j)^2 },
\label{eq:residue}
\end{equation}
where $|I|$ is the total number of rows of the bicluster, $|J|$ is the total number of columns, and $r(i,j)$ is given by
\begin{equation}
r(i,j) = \mathbf{B}(i,j) - \mathbf{B}(I,j) - \mathbf{B}(i,J) + \mathbf{B}(I,J).
\label{eq:res}
\end{equation}

Due to the presence of noise in most real-world data, the minimization of the MSR score should be constrained by a threshold $\delta$~\cite{cheng2000}. Another validation measure is the size of the bicluster, often called as its {\em volume}~\cite{defranca2013}, since usually large bicluster modules are sought. However, a good balance between coherence and volume is necessary, since they tend to be conflictive in nature. Moreover, the calibration of coherence and volume thresholds may not be trivial due to the different characteristics of each data set.  

Huang et al.~\cite{huang2011} devised a new additive coherence-based biclustering algorithm as part of an unsupervised feature ranking approach. An interesting property of this algorithm is that it allows the elicitation of {\em overlapped} biclusters, that is, submatrices sharing rows and/or columns. Moreover, it is based on two easy-to-apply (computationally efficient) procedures, namely: 1) the use of conventional (agglomerative with average linkage) hierarchical clustering of instances for each feature; and 2) the local search for biclusters based on operations of expansion, refinement, and merging from the set of clusters discovered in the first step, which are cast as bicluster seeds. Besides, such algorithm has only two hyperparameters to be calibrated, $T_d$ and $T_m$, which are respectively related to the volume and coherence of the resulting biclusters. While $T_m$ has the same role of Cheng and Church's $\delta$ \cite{cheng2000}, $T_d$ specifies the level at which the dendrogram associated with each feature should be cut and, thus, delimits the size of the resulting bicluster seeds. Once the data of each column is normalized via a standardization procedure, the sensitivity of the algorithm's performance to the calibration of these parameters should not be high~\cite{huang2011}.

Recently, coherence biclustering has been also adop\-ted as a source of accuracy improvement in the supervised classification of objects. For example, de França et al.~\cite{prati2013,defranca2015} have proposed novel algorithms for coping with multilabel classification and classification with noisy labels, respectively, by replacing or augmenting the feature space through the elicitation of good discriminative biclusters. By this means, novel binary features are extracted, each representing a discriminative bicluster between two classes of instances, and novel instances can be classified according to the way they match to these local patterns.

By other means, Odibat et al.~\cite{odibat2010} proposed the {\em DiBiCLUS} algorithm, aiming to mine differential biclusters from gene expression data labeled according to the types of experimental conditions (i.e., classes of features). In this context, for each gene, over-expressed (under-expressed) conditions are represented by positive (negative) numbers, and thus two genes are regarded as positively (negatively) co-expressed if they have the same (different) signs in a subset of conditions. The biclusters produced by {\em DiBiCLUS} have genes much co-expressed (correlated) in one class of conditions, but not in the other, or may have different types of co-expression among the two classes. This notion of differential biclusters has also been considered in the work of Wang et al.~\cite{wang2013}, who proposed a more efficient algorithm, referred to as {\em DECluster}. Such algorithm is based on the construction of a weighted undirected graph among the experimental conditions over which the differential biclusters are induced.

The {\em SDC} algorithm~\cite{fang2010}, in turn, is based on a frequent pattern mining algorithm (viz., Apriori) and seeks for patterns for which the difference between the correlations in the two classes of conditions is above a fixed threshold. On the other hand, more recently, Odibat and Reddy~\cite{odibat2014} proposed a novel algorithm, referred to as {\em Di-RAPOCC} (after Discriminative Ranking-based Arbitrarily Positioned Overlapping Co-clustering), which can extract large and arbitrarily-positioned biclusters containing both positively and negatively correlated objects but showing low inter-class overlap.

Contrary to the abovementioned works, this study focuses specifically on the improvements in accuracy/ge\-neralization that discriminative biclusters can bring to simple classification models such as perceptrons. Instead of considering only bioinformatics problems, as the majority of the reviewed works do, we have assessed the potentials of bicluster-enhanced perceptrons on a range of classification domains. Moreover, our focus is on the classification of data instances and not on the discrimination of classes related to the features (like the experimental conditions on bioinformatics data). More importantly, the algorithm used in BicNeuron for eliciting discriminative biclusters is conceptually different in the sense that it is based on contrastive data groups coming in pairs~\cite{dong2012}. Finally, it is straightforward and simple to implement, contrary to other approaches, such as {\em Di-RAPOCC}, which involve some intricate steps.

\section{Inducing perceptrons from discriminative biclusters}
\label{sec:bicneuron}

From BicNeuron's perspective, a high-quality bicluster denotes a subset of data instances from a given class that shows not only high coherence (i.e., well alignment) across a subset of features but also high separation from the nearest instances of the opposite class, when the latter are also projected onto the same feature subset. Such interpretation entails a pair of biclusters (one for each class) representing a particular data subspace where there is a salient contrast between the classes~\cite{dong2012}. Hence, we refer to these groups as {\em contrastive biclusters}. 

The strategy followed by BicNeuron is first to induce a series of coherent biclusters from only one of the classes (for this purpose, we have adopted the algorithm proposed by Huang et al.~\cite{huang2011}) and then to {\em artificially} generate biclusters composed of instances from the other class to become associated with the former. Hopefully, the new artificially-generated biclusters will be {\em non-coherent} in nature in a manner as to yield a good separation region between the classes. In fact, if both biclusters of a pair were highly coherent, their associated instances would be well aligned to each other, hindering their linear discrimination. From the resulting set of candidate pairs of biclusters, we filter out those that really allow a considerable contrast between the classes, as quantified by a proper measure. Upon each local subspace associated with a pair of good contrastive biclusters a perceptron model is trained, and the final classifier to predict the label of new data instances is chosen according to a model selection criterion.

\begin{figure*}
\scriptsize
\setlength{\tabcolsep}{2.5pt}	
\renewcommand{\arraystretch}{0.9}
\subfigure[]{
\begin{tabular}{|c||ccccc|}
\hline
 & $\mathbf{f}_1$ & $\mathbf{f}_2$ & $\mathbf{f}_3$ & $\mathbf{f}_4$ & $L$ \\
\hline\hline
$\mathbf{o}_1$ &  $\mathbf{15}$ & $\mathbf{12}$ & \multicolumn{1}{c|}{$\mathbf{20}$} & 13 & -1\\
$\mathbf{o}_2$ & $\mathbf{20}$ & $\mathbf{18}$ & \multicolumn{1}{c|}{$\mathbf{25}$} & 07 &-1\\
\cline{4-5}
$\mathbf{o}_3$ & $\mathbf{25}$ & $\mathbf{22}$ & \multicolumn{1}{|c|}{$\mathbf{30}$} &\multicolumn{1}{c|}{$\mathbf{35}$} &-1\\
\cline{2-4}
$\mathbf{o}_4$ & 33 & 07 & \multicolumn{1}{|c}{$\mathbf{46}$} & \multicolumn{1}{c|}{$\mathbf{50}$} &-1\\
\cline{4-5}
$\mathbf{o}_5$ & 10 & 10 & 35 & 45 & +1\\
$\mathbf{o}_6$ & 15 & 30 & 40 & 41 & +1\\
$\mathbf{o}_7$ & 22 & 20 & 20 & 10 & +1\\
$\mathbf{o}_8$ & 30 & 15 & 25 & 32 & +1\\
$\mathbf{o}_9$ & 20 & 17 & 30 & 50 & +1\\
\hline
\end{tabular}
}\hspace{.1cm}
\subfigure[]{
\begin{tabular}{|c||ccccc|}
\hline
 & $\mathbf{f}_1$ & $\mathbf{f}_2$ & $\mathbf{f}_3$ & $\mathbf{f}_4$ & $L$ \\
\hline\hline
$\mathbf{o}_1$ & $\mathbf{0.22}$ & $\mathbf{0.22}$  & \multicolumn{1}{c|}{$\mathbf{0.00}$} &  0.14 & -1\\
$\mathbf{o}_2$ & $\mathbf{0.43}$  & $\mathbf{0.48}$  & \multicolumn{1}{c|}{$\mathbf{0.19}$}  & 0.00   &-1\\
\cline{4-5}
$\mathbf{o}_3$ & $\mathbf{0.65}$  & $\mathbf{0.65}$ & \multicolumn{1}{|c|}{$\mathbf{0.38}$}  & \multicolumn{1}{c|}{$\mathbf{0.65}$} &-1\\
\cline{2-4}
$\mathbf{o}_4$ & 1.00  & 0.00 & \multicolumn{1}{|c}{$\mathbf{1.00}$}  & \multicolumn{1}{c|}{$\mathbf{1.00}$}  &-1\\
\cline{4-5}
$\mathbf{o}_5$ & 0.00  & 0.13 & 0.58  & 0.88 & +1\\
$\mathbf{o}_6$ & 0.22 & 1.00  & 0.77 & 0.79 & +1\\
$\mathbf{o}_7$ & 0.52 & 0.57 & 0.00  &  0.07 & +1\\
$\mathbf{o}_8$ & 0.87  & 0.35 & 0.19  & 0.58 & +1\\
$\mathbf{o}_9$ & 0.43 & 0.43 & 0.38  & 1.00  & +1\\
\hline
\end{tabular}
}\hspace{.1cm}
\subfigure[]{
\begin{minipage}{.15\linewidth}
\begin{tabular}{|c||ccc|}
\hline
& $\mathbf{f}_1$ & $\mathbf{f}_2$ & $\mathbf{f}_3$ \\
\hline\hline
$\mathbf{o}_1$ & 0.22 & 0.22  & 0.00 \\
$\mathbf{o}_2$ & 0.43  & 0.48  & 0.19 \\
$\mathbf{o}_3$ & 0.65  & 0.65 & 0.38  \\
\hline
$\mathbf{c}_1$ & $\mathbf{0.43}$ & $\mathbf{0.45}$  &   $\mathbf{0.19}$\\
\hline
\end{tabular}

\vspace{.1cm}
\begin{tabular}{|c||cc|}
\hline
& $\mathbf{f}_3$ & $\mathbf{f}_4$\\
\hline\hline
$\mathbf{o}_3$ & 0.38  & 0.65\\
$\mathbf{o}_4$ & 1.00  & 1.00\\
\hline
$\mathbf{c}_2$ & $\mathbf{0.69}$ & $\mathbf{0.82}$\\
\hline
\end{tabular}
\end{minipage}
}\hspace{.17cm}
\subfigure[]{
\begin{minipage}{.15\linewidth}
\begin{tabular}{|c||c|c|c|c|c|}
\hline
& $\mathbf{o}_5$ & $\mathbf{o}_6$ & $\mathbf{o}_7$ & $\mathbf{o}_8$ & $\mathbf{o}_9$\\
\hline\hline
$\mathbf{c}_1$ & 0.67 & 0.83 & $\mathbf{0.24}$ & $\mathbf{0.45}$ & $\mathbf{0.19}$\\
\hline
$\mathbf{c}_2$ & $\mathbf{0.12}$ & $\mathbf{0.09}$ & 1.02 & 0.56 & 0.36\\
\hline
\end{tabular}

\vspace{.1cm}
\begin{tabular}{|c||ccc|}
\hline
& $\mathbf{f}_1$ & $\mathbf{f}_2$ & $\mathbf{f}_3$ \\
\hline\hline
$\mathbf{o}_7$ & 0.52 & 0.57 & 0.00  \\
$\mathbf{o}_8$ & 0.87  & 0.35 & 0.19 \\
$\mathbf{o}_9$ & 0.43 & 0.43 & 0.38  \\
\hline
\end{tabular}

\vspace{.15cm}
\begin{tabular}{|c||cc|}
\hline
& $\mathbf{f}_3$ & $\mathbf{f}_4$\\
\hline\hline
$\mathbf{o}_5$ & 0.58  & 0.88 \\
$\mathbf{o}_6$ & 0.77 & 0.79 \\
\hline
\end{tabular}
\end{minipage}
}\hspace{.1cm}
\subfigure[]{
\begin{minipage}{.15\linewidth}
\vspace{1.3cm}
\begin{tabular}{|c||c|c|}
\hline
& $\mathbf{B}_1$ & $\mathbf{B}^c_1$\\
\hline\hline
$MSR$ & 0.0002 & 0.0209\\
\hline\hline
& $\mathbf{B}_2$ & $\mathbf{B}^c_2$\\
\hline\hline
$MSR$ & 0.0045 & 0.0049\\
\hline\hline
& $\mathbf{B}_1/\mathbf{B}^c_1$ & $\mathbf{B}_2/\mathbf{B}^c_2$\\
\hline\hline
$r_{MSR}$& $\mathbf{0.01}$ & 0.93\\
\hline
\end{tabular}
\end{minipage}
}
\caption{First four steps of BicNeuron: (a) Original data set with nine objects $\mathbf{o}_i$, four features $\mathbf{f}_j$, and two labels $L=\{-1,+1\}$; (b) Normalized data, showing two coherent biclusters ($\mathbf{B}_1$ and $\mathbf{B}_2$) for the first (minor) class; (c) Centroids ($\mathbf{c}_1$ and $\mathbf{c}_2$) of each bicluster; (d) Based on the distances between the objects of the second class and the bicluster centroids (upper table), the contrastive biclusters ($\mathbf{B}^c_1$ and $\mathbf{B}^c_2$) are generated (two lower tables, respectively); and (e) Assuming a threshold $\tau \leq 0.9$ and considering the MSR ratio for each pair of contrastive biclusters, only the first pair should be regarded as a good discriminative subspace for training the final perceptron.}
\label{fig:contbics}
\end{figure*}

In a more formal manner, BicNeuron involves the following six steps:
\begin{enumerate}
\item \textbf{Uncovering coherent biclusters from one of the classes:} Firstly, the binary training data set $\mathbf{X}=[R,C]$ is normalized (via max-min normalization) and the instances are divided into two subsets, $\mathbf{X}_1=[R_1,C]$ and $\mathbf{X}_2=[R_2,C]$, according to their class labels. Then, a biclustering algorithm is applied, in an unsupervised manner, solely to $\mathbf{X}_1$ (by default, the class with less instances), yielding a set of coherent biclusters $\mathcal{B}$.
\item \textbf{Computing the bicluster centroids:} For each bicluster $\mathbf{B} = [I \subset R_1,J \subset C] \in \mathcal{B}$, a new average pattern (row centroid) $\mathbf{c_B}$ is generated as 
\begin{equation}
\mathbf{c_B} = \frac{1}{|I|}\sum_{i \in I} \mathbf{B}(i,J).
\end{equation}
\item \textbf{Generating the contrastive biclusters:} For each bicluster $\mathbf{B} = [I,J]\in \mathcal{B}$, a contrastive bicluster $\mathbf{B}^c = [I^c \subset R_2,J]$ is artificially created such that $I^c$ comprises the nearest objects from the other class, as measured by their Euclidean distances to $\mathbf{c_B}$, considering only the features $J$. It is important to stress again that $\mathbf{B}^c$ should not be as coherent as $\mathbf{B}$. In fact, the less coherent $\mathbf{B}$ is, the better would be the contrast between the pair and, thus, between the classes. A centroid $\mathbf{c_{B^c}}$ may then be computed for each new bicluster $\mathbf{B}^c$.
This step results in a set of contrastive biclusters $\mathcal{B}^c$, such that $|\mathcal{B}| = |\mathcal{B}^c|$.
\item \textbf{Filtering the contrastive biclusters:} The pairs of contrastive biclusters $\{(\mathbf{B},\mathbf{B}^c),$ $\mathbf{B} \in \mathcal{B} , \mathbf{B}^c \in \mathcal{B}^c\}$ are sorted according to their levels of contrast, and then only those pairs with high discrimination are kept. Here, we adopt as ranking criterion the {\em coherence contrast} between the contrastive biclusters, meaning that pairs showing high differentiation between their MSR values are preferred. To fulfill this criterion, the {\em MSR ratio} measure was used, which, for the pair $(\mathbf{B}=[I,J],\mathbf{B}^c=[I^c,J])$, is given as: 
\begin{equation} 
r_{MSR}(\mathbf{B},\mathbf{B}^c) = \frac{MSR(\mathbf{B})}{MSR(\mathbf{B}^c)} = \frac{H(I,J)}{H(I^c,J)},
\end{equation}
where $MSR(\mathbf{B})=H(I,J)$ denotes the coherence value of $\mathbf{B}$ as calculated by \eqref{eq:residue}. Since $MSR(\mathbf{B}^c) > MSR(\mathbf{B})$,  $r_{MSR}(\mathbf{B},\mathbf{B}^c)$ will never be undefined (i.e., $\frac{0}{0}$) and will only be zero (optimal value) when $\mathbf{B}$ is a perfect coherent bicluster. After ranking the pairs of contrastive biclusters by the values of their MSR ratio, only those for which $r_{MSR} \leq \tau$, with $\tau$ being a previously fixed threshold, will be kept for the next step. We denote as $N_B$ the number of selected pairs of contrastive biclusters, and as $\mathcal{B}_s$ and $\mathcal{B}_s^c$ the sets of selected biclusters from the two classes, respectively, such that $N_B = |\mathcal{B}_s|=|\mathcal{B}_s^c|$.
\item \textbf{Inducing the perceptron models:} From each pair $(\mathbf{B} = [I,J],\mathbf{B}^c = [I^c,J])$, with $\mathbf{B} \in \mathcal{B}_s$ and $\mathbf{B}^c \in \mathcal{B}_s^c$, a new data set $\mathbf{X}_{(\mathbf{B}, \mathbf{B}^c)}$ is assembled by concatenating the instances $I$ and $I^c$ projected onto $J$, and then a (standard or kernel) perceptron model $p$ is trained on $\mathbf{X}_{(\mathbf{B}, \mathbf{B}^c)}$. This results in a set of  perceptron models $\mathcal{P} = \{p_k\}, k =1,\ldots,N_B$.
\item \textbf{Selecting the final classifier:} From $\mathcal{P}$, the best model is selected according to a given performance measure. Here, we adopt the area under the receiver operating characteristic curve (AUC), since this measure can provide an unbiased estimate of accuracy averaged over different loss conditions~\cite{alpaydin2014}. Moreover, it yields a good assessment of the accuracies delivered for different classes when the data set is imbalanced. The perceptron $p^* \in \mathcal{P}$ with highest AUC value calculated over the whole training set $\mathbf{X}$ is returned as the final classifier.
\end{enumerate}

Figure~\ref{fig:contbics} brings an example to illustrate how contrastive biclusters are produced and selected in BicNeuron. In principle, any biclustering algorithm could be employed to evoke the initial set of coherent biclusters from the first class. However, due to the satisfactory properties reviewed in Section~\ref{sec:bics}, we have made use of the algorithm conceived by Huang et al.~\cite{huang2011} while conducting the experiments reported in the next section. 

Considering a training data set $\mathbf{X}=[R,C]$, with $N=|R|$ and $M=|C|$, the time complexity of Huang et al.'s algorithm is bounded by $O(N(N+M)^2)$, which is also the complexity of the first step of BicNeuron -- notice that, in the worst case, each class has the same number of instances. Considering now that the time costs entailed by the operations over the contrastive biclusters is bounded by $O(|\mathcal{B}|NM)$, where $|\mathcal{B}|$ is the number of biclusters delivered in the first step, one can assert that the time complexity of the first four steps of BicNeuron is limited by $O(N(N+M)^2+|\mathcal{B}|NM)$. The time costs due to the last two steps depend on exogeneous variables related to the training of the perceptrons and the calculation of their AUC values over $\mathbf{X}$, but since these operations are usually not costly if the number of training epochs is fixed to a reasonable value we can state that the computational performance of BicNeuron is mostly influenced by the biclustering algorithm in use.

In what concerns the possibility of extracting coherent biclusters from the data of the second (largest) class, this could in principle be also pursued. However, one should notice that this would significantly increase the time costs of the whole approach since the biclustering algorithm (actually all BicNeuron steps but the last one) would run twice. Concerning more specifically the size (in terms of number of instances) of the contrastive biclusters generated in the third step, in theory this value could be somehow tuned based on the characteristics of the classification problem in sight. However, for the sake of simplicity, we have assumed that $|I| = |I^c|$ for each pair of contrastive biclusters $(\mathbf{B} = [I,J],\mathbf{B}^c = [I^c,J])$. On the other hand, different criteria and measures could be adopted in the fourth and sixth steps for the purpose of quantifying the contrast between the biclusters and selecting the best perceptron model, respectively. For instance, the {\em geometrical separation} between the contrastive biclusters could be used as an alternative criterion to the coherence contrast. In this case, different notions of {\em margin} could be adopted as measures for ranking the pairs $(\mathbf{B},\mathbf{B}^c)$, such as taking the Euclidean distance between their centroids $\mathbf{c_B}$ and $\mathbf{c_{B^c}}$ or between their farthest instances. Yet, the analysis of the impact of such alternative criteria/measures on the accuracy performance of BicNeuron is out of the scope of this paper.

\section{Computational experiments}
\label{sec:exps}

In order to assess the potentials of BicNeuron in leveraging the classification performance displayed by perceptrons, we have developed a prototype in Python based on the Scikit-Learn toolkit (\url{http://scikit-learn.org/stable/}) and conducted an extensive series of experiments. In the next subsections, we give details on the experimental setup and then present and discuss the main results achieved.

\subsection{Experimental setup}
\label{sec:setup}

Thirteen binary classification data sets from publicly-available repositories~\cite{lichman2013,delve1998,desouto2008,costa2013} were used for the purpose of assessing the quality of the linear subspaces uncovered by BicNeuron via analysis of the performance of the associated perceptrons. Besides, one multiclass data set related to the difficult task of epilepsy diagnosis based on electroencephalogram (EEG) signal analysis~\cite{lima2011,nunes2014} was also adopted, yielding 10 additional binary classification problems. For the latter data set, feature extraction was performed over the raw EEG signals via discrete wavelet transform~\cite{tang2009} using Daubecchies of order 4 as wavelet function. Such decision complies with previous work on the subject~\cite{subasi2008,lima2011}.

The choice of these data sets was mostly motivated by their distinct natures (they stem from different application domains) and structural properties, such as the number of instances, number of features, and class distributions (Table~\ref{tab:datasets}). Besides, Duch et al. \cite{duch2012} showed experimentally that some of them should be regarded as `non-trivial' for simple classifiers. All of these data sets comprise real-valued features only, which is a restriction imposed by the biclustering algorithm used. Moreover, we have focused on data sets with no missing values and moderate/large numbers of features, since it would be meaningless to extract biclusters from low dimensional spaces.

\begin{table*}
\caption{Characterization of the data sets.}
\centering
\setlength{\tabcolsep}{2.5pt}
\renewcommand\arraystretch{.9}
\begin{tabular}{ccccc}
\hline
Index & Data set & Instances & Features & Class Dist.  \\
\hline
1 & chen & 179 & 85 & (41.9\%;58.1\%)\\
2 & chowdary & 104 & 182 & (40.4\%;59.6\%)\\
3 & colon & 62 & 2000 & (35.5\%;64.5\%)\\
4 & ionosphere & 351 & 32 & (35.9\%;64.1\%)\\
5 & lsvt & 126 & 310 & (33.3\%;66.7\%)\\
6 & parkinsons & 195 & 22 & (24.6\%;75.4\%)\\
7 & ringnorm & 7400 & 20 & (49.5\%;50.5\%)\\
8 & singh & 102 & 339 & (49.0\%;51.0\%)\\
9 & sonar & 208 & 61 & (46.6\%;53.4\%)\\
10 & spambase & 4601 & 57 & (39.4\%;60.6\%)\\
11 & tipspam & 2762 & 23 & (50.0\%;50.0\%)\\
12 & twonorm & 7400 & 20 & (50.0\%;50.0\%)\\
13 & wdbc & 569& 31 & (37.3\%;62.7\%)\\
14 & eeg & 500 & 40 & (20\%;20\%;20\%;20\%;20\%)\\
\hline
\end{tabular}
\label{tab:datasets}
\end{table*}

In its present version, BicNeuron has three control parameters to be calibrated, namely: $T_d$ and $T_m$, associated with the biclustering algorithm; and $\tau$, the threshold used for selecting the pairs of contrastive biclusters. In our experiments, we have kept $T_m$ fixed in 0.02, as suggested by~Huang et al.~\cite{huang2011}, but varied systematically the value of $T_d$ in the range $[0.5, 0.8, 1.0, 1.5]$. (For the {\em ionosphere} data set, in particular, since no pair of contrastive biclusters was found for these settings, the value of $T_d$ was set as $[0.005, 0.01, 0.05]$.) Conversely, $\tau$ has assumed values in the range $[0.1, 0.3, 0.5, 0.7, 0.9]$ in order to capture different notions of strictness for the criterion of coherence contrast.

For each data set, the assessment was realized via 10-fold stratified cross-validation~\cite{alpaydin2014}. For each of the 10 iterations, the control parameter values of BicNeuron were also calibrated via cross-validation, but performed solely on the training partition.  Standard perceptron (SP) and kernel perceptron (KP) models were also trai\-ned and tested on the same folds as BicNeuron to serve as baseline for comparison. Contrary to KP, the training of SP was performed via stochastic gradient descent (online mode) with data being shuffled at each epoch. Besides, both the linear kernel with no bias term (i.e. $K_L(\mathbf{x}_l,\mathbf{x}_k) = \mathbf{x}_l \cdot \mathbf{x}_k$) and RBF kernel (i.e. $K_{RBF}(\mathbf{x}_l,\mathbf{x}_k) = \exp(-\frac{||\mathbf{x}_l - \mathbf{x}_k||^2}{2\sigma^2}$)) were adopted for KP. After preliminary experimentation, the number of training epochs for both SP and KP were set at 20, whereas 0.1 was adopted as the value of learning rate ($\beta$) for SP and also as the value of $\sigma$ for the RBF kernel. These settings were also used for the perceptron models induced with BicNeuron.  

For the classifier generated by the best BicNeuron configuration of each cross-validation iteration as well as for SP and KP, results were collected for each of the following measures: 1) overall accuracy (ACC); 2) accuracy for the minority class (ACCm); 3) accuracy for the majority class (ACCM); and 4) AUC. Finally, the non-parametric Wilcoxon test~\cite{demsar2006} was applied separately to the ACC/AUC results achieved for each data set in order to check whether the difference in performance between the classifiers was significant or not. 

\begin{table*}
\small
\centering
\setlength{\tabcolsep}{1pt}	
\caption{ACC results achieved by the linear classifiers for each binary classification data set. Performance is measured separately for each test fold of the cross-validation process as well as in terms of average ($\mu \pm \sigma$) and best values. Cases highlighted in bold (italics) are those in which the performance of BicNeuron is significantly better (worse) than that of the given contestant as measured by the Wilcoxon test ($p$-value $\leq 0.01$).  Best average results are underlined for each data set.}
\begin{tabular}{cccccccccccccc}
\hline
& & \multicolumn{12}{c}{ACC}\\
\cline{3-14}
Model & Data set  & 1 &   2 &     3 &     4 &     5 &     6 &     7 &     8 &     9 &     10 & $\mu \pm \sigma$ & Best \\
\hline
SP & \multirow{3}{*}{chen} & 0.94&0.68&0.74&0.95&0.94&0.94&0.82&0.94&0.71&1.00 & 0.87 $\pm$ 0.12 & 1.00\\
KP & & 1.00&0.63&0.63&0.79&0.56&0.76&0.71&0.82&0.65&0.68 & {\bf 0.72 $\pm$ 0.13} & 1.00 \\
BicNeuron &  & 1.00&0.79&0.79&0.95&1.00&1.00&0.88&0.88&0.94&0.95 & \underline{0.92 $\pm$ 0.08} & 1.00 \\
\hline
SP & \multirow{3}{*}{chowdary} & 0.90&1.00&1.00&1.00&0.90&1.00&1.00&0.90&1.00&0.92 & 0.96 $\pm$ 0.05 & 1.00 \\
KP & & 1.00&1.00&1.00&1.00&1.00&1.00&1.00&0.80&1.00&0.83 & 0.96 $\pm$ 0.08 & 1.00\\
BicNeuron &  & 1.00&1.00&1.00&1.00&1.00&1.00&1.00&0.90&1.00&0.92 & \underline{0.98 $\pm$ 0.04} & 1.00 \\
\hline
SP & \multirow{3}{*}{colon} & 0.83&0.71&0.83&1.00&1.00&0.83&0.67&0.67&0.50&1.00 & \underline{0.80 $\pm$ 0.17} & 1.00 \\
KP & & 0.83&0.57&0.67&0.67&0.67&0.83&0.83&0.67&0.50&0.57 & 0.68 $\pm$ 0.12 & 0.83\\
BicNeuron &  & 0.83&0.86&0.33&0.83&0.83&1.00&1.00&0.50&0.50&0.86 & 0.75 $\pm$ 0.23 & 1.00\\
\hline
SP & \multirow{3}{*}{ionosphere} & 0.88&0.67&0.78&0.75&0.86&0.86&0.88&0.88&0.82&0.78 & \underline{0.82 $\pm$ 0.07} & 0.88 \\
KP & & 0.65&0.64&0.64&0.64&0.64&0.63&0.65&0.65&0.65&0.67 & {\bf 0.64 $\pm$ 0.01} & 0.67 \\
BicNeuron &  & 0.82&0.78&0.81&0.72&0.75&0.66&0.79&0.79&0.74&0.69 & 0.76 $\pm$ 0.05 & 0.82  \\
\hline
SP & \multirow{3}{*}{lsvt} & 0.67&1.00&0.92&0.92&0.77&0.92&0.75&0.83&0.83&0.92 & 0.85 $\pm$ 0.10 & 1.00 \\
KP & & 0.67&0.69&0.69&0.69&0.54&0.62&0.75&0.67&0.67&0.69 & {\bf 0.67 $\pm$ 0.06} & 0.75 \\
BicNeuron &  & 0.75&0.92&0.92&0.92&0.77&0.92&0.83&0.92&0.92&0.85 & \underline{0.87 $\pm$ 0.07} & 0.92  \\
\hline
SP & \multirow{3}{*}{parkinsons} & 0.61&0.45&0.60&0.50&0.75&0.45&0.05&0.74&0.72&0.70 & {\bf 0.56 $\pm$ 0.21} & 0.75\\
KP & & 0.78&0.95&0.75&0.75&0.80&0.90&0.75&0.74&0.78&1.00 & 0.82 $\pm$ 0.10 & 1.00 \\
BicNeuron &  & 0.78&1.00&0.85&0.85&0.75&0.90&0.75&0.79&0.89&0.95 & \underline{0.85 $\pm$ 0.09} & 1.00  \\
\hline
SP & \multirow{3}{*}{ringnorm} & 0.50&0.52&0.49&0.69&0.73&0.50&0.50&0.49&0.50&0.54 & \underline{0.55 $\pm$ 0.09} & 0.73 \\
KP & & 0.50&0.50&0.50&0.50&0.51&0.51&0.50&0.50&0.50&0.50 & {\bf 0.50 $\pm$ 0.00} & 0.51 \\
BicNeuron &  & 0.52&0.57&0.53&0.57&0.56&0.56&0.54&0.54&0.57&0.50 & \underline{0.55 $\pm$ 0.02} & 0.57  \\
\hline
SP & \multirow{3}{*}{singh} & 0.50&0.73&0.80&0.90&0.60&0.60&0.60&0.50&0.70&0.64 & {\bf 0.66 $\pm$ 0.13} & 0.90 \\
KP & & 0.50&0.55&0.70&0.80&0.50&0.80&0.60&0.70&0.60&0.64 & {\bf 0.64 $\pm$ 0.11} & 0.80 \\
BicNeuron &  & 0.80&1.00&0.90&0.80&1.00&0.80&0.80&0.70&0.80&0.91 & \underline{0.85 $\pm$ 0.10} & 1.00  \\
\hline
SP & \multirow{3}{*}{sonar} & 0.55&0.67&0.52&0.57&0.52&0.52&0.52&0.55&0.55&0.77 & {\bf 0.58 $\pm$ 0.08} & 0.77 \\
KP & & 0.55&0.90&0.67&0.81&0.52&0.76&0.71&0.60&0.55&0.95 & 0.70 $\pm$ 0.15 & 0.95 \\
BicNeuron &  & 0.85&0.86&0.76&0.90&0.86&0.81&0.76&1.00&0.60&0.73 & \underline{0.81 $\pm$ 0.11} & 1.00  \\
\hline
SP & \multirow{3}{*}{spambase} & 0.61&0.61&0.61&0.61&0.61&0.61&0.61&0.61&0.61&0.61 & {\bf 0.61 $\pm$ 0.00} & 0.61 \\
KP & & 0.61&0.60&0.63&0.62&0.60&0.61&0.63&0.61&0.64&0.59 & {\bf 0.61 $\pm$ 0.01} & 0.64 \\
BicNeuron &  & 0.70&0.68&0.71&0.71&0.63&0.70&0.69&0.68&0.69&0.70 & \underline{0.69 $\pm$ 0.02} & 0.71  \\
\hline
SP & \multirow{3}{*}{tipspam} & 0.50&0.50&0.50&0.50&0.50&0.50&0.52&0.54&0.50&0.50 & {\bf 0.51 $\pm$ 0.01} & 0.54 \\
KP & & 0.49&0.50&0.49&0.50&0.50&0.50&0.50&0.50&0.50&0.50 & {\bf 0.50 $\pm$ 0.00} & 0.50 \\
BicNeuron &  & 0.71&0.70&0.75&0.65&0.59&0.69&0.75&0.73&0.73&0.67 & \underline{0.70 $\pm$ 0.05} & 0.75 \\
\hline
SP & \multirow{3}{*}{twonorm} & 0.74&0.87&0.80&0.75&0.77&0.89&0.77&0.89&0.92&0.62 & \underline{{\it 0.80 $\pm$ 0.09}} & 0.92 \\
KP & & 0.50&0.50&0.50&0.50&0.50&0.50&0.50&0.50&0.50&0.50 & {\bf 0.50 $\pm$ 0.00} & 0.50 \\
BicNeuron &  & 0.60&0.60&0.62&0.58&0.63&0.60&0.59&0.61&0.62&0.62 & 0.61 $\pm$ 0.01 & 0.63  \\
\hline
SP & \multirow{3}{*}{wdbc} & 0.95&0.97&0.91&0.86&0.96&0.93&0.93&0.91&0.98&0.90 & 0.93 $\pm$ 0.04 & 0.98 \\
KP & & 0.66&0.62&0.63&0.63&0.65&0.75&0.67&0.70&0.70&0.78 & {\bf 0.68 $\pm$ 0.05} & 0.78 \\
BicNeuron &  & 0.96&0.90&0.89&0.95&0.98&0.98&0.93&0.98&0.93&0.93 & \underline{0.94 $\pm$ 0.03} & 0.98  \\
\hline
\end{tabular}
\label{tab:acc}
\end{table*}

\begin{table*}
\small
\centering
\setlength{\tabcolsep}{1pt}	
\caption{AUC results achieved by the linear classifiers for each binary classification data set. Performance is measured separately for each test fold of the cross-validation process as well as in terms of average ($\mu \pm \sigma$) and best values. Cases highlighted in bold (italics) are those in which the performance of BicNeuron is significantly better (worse) than that of the given contestant as measured by the Wilcoxon test ($p$-value $\leq 0.01$). Best average results are underlined for each data set.}
\begin{tabular}{cccccccccccccc}
\hline
& & \multicolumn{12}{c}{AUC}\\
\cline{3-14} 
Model & Data set  & 1 &     2 &     3 &     4 &     5 &     6 &     7 &     8 &     9 &     10 & $\mu \pm \sigma$ & Best \\
\hline
SP & \multirow{3}{*}{chen} & 0.95&0.62&0.69&0.95&0.94&0.93&0.79&0.95&0.64&1.00 & 0.85 $\pm$ 0.15 & 1.00\\
KP & & 1.00&0.58&0.56&0.75&0.50&0.71&0.64&0.79&0.57&0.62 & {\bf 0.67 $\pm$ 0.15} & 1.00\\
BicNeuron &  & 1.00&0.75&0.75&0.95&1.00&1.00&0.90&0.88&0.93&0.94 & \underline{0.91 $\pm$ 0.09} & 1.00\\
\hline
SP & \multirow{3}{*}{chowdary} & 0.88&1.00&1.00&1.00&0.88&1.00&1.00&0.88&1.00&0.93 & 0.96 $\pm$ 0.06 & 1.00\\
KP & & 1.00&1.00&1.00&1.00&1.00&1.00&1.00&0.75&1.00&0.86 & 0.96 $\pm$ 0.09 & 1.00\\
BicNeuron &  & 1.00&1.00&1.00&1.00&1.00&1.00&1.00&0.88&1.00&0.93 & \underline{0.98 $\pm$ 0.04} & 1.00\\
\hline
SP & \multirow{3}{*}{colon} & 0.88&0.71&0.75&1.00&1.00&0.75&0.50&0.62&0.50&1.00 & \underline{0.77 $\pm$ 0.19} & 1.00\\
KP & & 0.75&0.50&0.50&0.50&0.50&0.75&0.75&0.50&0.50&0.50 & 0.57 $\pm$ 0.12 & 0.75\\
BicNeuron &  & 0.75&0.83&0.25&0.75&0.75&1.00&1.00&0.38&0.50&0.83 & 0.70 $\pm$ 0.25 & 1.00\\
\hline
SP & \multirow{3}{*}{ionosphere} &  0.83&0.54&0.69&0.65&0.81&0.81&0.83&0.83&0.75&0.69 & \underline{0.74 $\pm$ 0.10} & 0.83\\
KP & & 0.50&0.50&0.50&0.50&0.50&0.50&0.50&0.50&0.50&0.54 & {\bf 0.50 $\pm$ 0.01} & 0.54\\
BicNeuron &  & 0.75&0.69&0.73&0.63&0.65&0.54&0.71&0.71&0.62&0.58 & 0.66 $\pm$ 0.07 & 0.75\\
\hline
SP & \multirow{3}{*}{lsvt} & 0.50&1.00&0.88&0.88&0.74&0.90&0.62&0.75&0.75&0.88 & 0.79 $\pm$ 0.15 & 1.00\\
KP & & 0.50&0.50&0.50&0.50&0.44&0.50&0.62&0.50&0.50&0.50 & {\bf 0.51 $\pm$ 0.05} & 0.62\\
BicNeuron &  & 0.81&0.94&0.88&0.94&0.74&0.90&0.75&0.88&0.94&0.89 & \underline{0.87 $\pm$ 0.08} & 0.94\\
\hline
SP & \multirow{3}{*}{parkinsons} & 0.75&0.63&0.73&0.67&0.83&0.63&0.10&0.56&0.73&0.80 & 0.64 $\pm$ 0.21 & 0.83\\
KP & & 0.50&0.90&0.50&0.50&0.60&0.87&0.50&0.50&0.50&1.00 & 0.64 $\pm$ 0.20 & 1.00\\
BicNeuron &  & 0.50&1.00&0.70&0.70&0.50&0.80&0.50&0.60&0.75&0.90 & \underline{0.70 $\pm$ 0.17} & 1.00\\
\hline
SP & \multirow{3}{*}{ringnorm} & 0.50&0.53&0.50&0.69&0.72&0.50&0.51&0.50&0.50&0.53 & \underline{0.55 $\pm$ 0.09} & 0.72\\
KP & & 0.50&0.50&0.50&0.50&0.50&0.50&0.50&0.50&0.50&0.50 & {\bf 0.50 $\pm$ 0.00} & 0.50\\
BicNeuron &  & 0.52&0.56&0.53&0.57&0.56&0.56&0.54&0.54&0.57&0.50 & \underline{0.55 $\pm$ 0.02} & 0.57\\
\hline
SP & \multirow{3}{*}{singh} & 0.50&0.70&0.80&0.90&0.60&0.60&0.60&0.50&0.70&0.60 & {\bf 0.65 $\pm$ 0.13} & 0.90\\
KP & & 0.50&0.50&0.70&0.80&0.50&0.80&0.60&0.70&0.60&0.60 & {\bf 0.63 $\pm$ 0.12} & 0.80\\
BicNeuron &  & 0.80&1.00&0.90&0.80&1.00&0.80&0.80&0.70&0.80&0.90 & \underline{0.85 $\pm$ 0.10} & 1.00\\
\hline
SP & \multirow{3}{*}{sonar} & 0.50&0.65&0.50&0.57&0.50&0.50&0.50&0.50&0.50&0.76 & {\bf 0.55 $\pm$ 0.09} & 0.76\\
KP & & 0.50&0.90&0.65&0.80&0.50&0.77&0.70&0.56&0.50&0.95 & 0.68 $\pm$ 0.17 & 0.95\\
BicNeuron &  & 0.83&0.85&0.76&0.90&0.86&0.81&0.75&1.00&0.59&0.75 & \underline{0.81 $\pm$ 0.11} & 1.00\\
\hline
SP & \multirow{3}{*}{spambase} & 0.50&0.50&0.50&0.50&0.50&0.50&0.50&0.50&0.50&0.50 & {\bf 0.50 $\pm$ 0.00} & 0.50\\
KP & & 0.51&0.50&0.53&0.51&0.50&0.51&0.53&0.50&0.54&0.49 & {\bf 0.51 $\pm$ 0.02} & 0.54\\
BicNeuron &  & 0.73&0.73&0.73&0.64&0.53&0.73&0.61&0.60&0.60&0.72 & \underline{0.66 $\pm$ 0.07} & 0.73\\
\hline
SP & \multirow{3}{*}{tipspam} &  0.50&0.50&0.50&0.50&0.50&0.50&0.52&0.54&0.50&0.50 & {\bf 0.51 $\pm$ 0.01} & 0.54\\
KP & & 0.49&0.50&0.49&0.50&0.50&0.50&0.50&0.50&0.50&0.50 & {\bf 0.50 $\pm$ 0.00} & 0.50\\
BicNeuron &  & 0.71&0.70&0.75&0.65&0.59&0.69&0.75&0.73&0.73&0.67 & \underline{0.70 $\pm$ 0.05} & 0.75\\
\hline
SP & \multirow{3}{*}{twonorm} &  0.74&0.87&0.80&0.75&0.77&0.89&0.77&0.89&0.92&0.61 & \underline{{\it 0.80 $\pm$ 0.09}} & 0.92\\
KP & &  0.50&0.50&0.50&0.50&0.50&0.50&0.50&0.50&0.50&0.50 & {\bf 0.50 $\pm$ 0.00} & 0.50\\
BicNeuron &  & 0.60&0.60&0.62&0.58&0.63&0.60&0.59&0.61&0.62&0.61 & 0.61 $\pm$ 0.01 & 0.63\\
\hline
SP & \multirow{3}{*}{wdbc} & 0.93&0.95&0.88&0.81&0.95&0.90&0.90&0.88&0.98&0.86 & 0.91 $\pm$ 0.05 & 0.98\\
KP & & 0.55&0.50&0.50&0.50&0.52&0.67&0.55&0.60&0.60&0.70 & {\bf 0.57 $\pm$ 0.07} & 0.70\\
BicNeuron &  & 0.96&0.89&0.88&0.94&0.99&0.98&0.90&0.99&0.91&0.94 & \underline{0.94 $\pm$ 0.04} & 0.99\\
\hline
\end{tabular}
\label{tab:auc}
\end{table*}

\begin{table*}
\caption{Accuracy results for the minor (ACCm) and major (ACCM) classes achieved by each classification approach for each binary classification data set. Performance is measured in terms of average ($\mu \pm \sigma$) and best values.}
\small
\centering
\setlength{\tabcolsep}{2.5pt}
\renewcommand\arraystretch{.9}
\begin{tabular}{cccccc}
\hline
& & \multicolumn{2}{c}{ACCm} & \multicolumn{2}{c}{ACCM} \\
Model & Data set & $\mu \pm \sigma$ & Best & $\mu \pm \sigma$ & Best \\
\hline
SP & \multirow{3}{*}{chen} & 0.72 $\pm$ 0.32 & 1.00 & 0.97 $\pm$ 0.05& 1.00\\
KP &  & 0.36 $\pm$ 0.29 & 1.00 & 0.99 $\pm$ 0.03 & 1.00 \\
BicNeuron &  & 0.86 $\pm$ 0.20 & 1.00 & 0.96 $\pm$ 0.07 & 1.00 \\
\hline
SP & \multirow{3}{*}{chowdary} & 0.93 $\pm$ 0.12 & 1.00 & 0.99 $\pm$ 0.05& 1.00\\
KP &  & 0.95 $\pm$ 0.16 & 1.00 & 0.97 $\pm$ 0.09 & 1.00 \\
BicNeuron &  & 0.97 $\pm$ 0.08 & 1.00 & 0.99 $\pm$ 0.05 & 1.00 \\
\hline
SP & \multirow{3}{*}{colon} & 0.67 $\pm$ 0.33 & 1.00 & 0.88 $\pm$ 0.18& 1.00\\
KP &  & 0.20 $\pm$ 0.26 & 0.50 & 0.95 $\pm$ 0.16 & 1.00 \\
BicNeuron &  & 0.53 $\pm$ 0.34 & 1.00 & 0.88 $\pm$ 0.21 & 1.00 \\
\hline
SP & \multirow{3}{*}{ionosphere} & 0.49 $\pm$ 0.20 & 0.67 & 1.00 $\pm$ 0.00& 1.00\\
KP &  & 0.01 $\pm$ 0.02 & 0.08 & 1.00 $\pm$ 0.00 & 1.00 \\
BicNeuron &  & 0.33 $\pm$ 0.14 & 0.50 & 1.00 $\pm$ 0.01 & 1.00 \\
\hline
SP & \multirow{3}{*}{lsvt} & 0.59 $\pm$ 0.29 & 1.00 & 0.99 $\pm$ 0.04& 1.00\\
KP &  & 0.03 $\pm$ 0.08 & 0.25 & 0.99 $\pm$ 0.04 & 1.00 \\
BicNeuron &  & 0.84 $\pm$ 0.19 & 1.00 & 0.89 $\pm$ 0.12 & 1.00 \\
\hline
SP & \multirow{3}{*}{parkinsons} & 0.82 $\pm$ 0.33 & 1.00 & 0.47 $\pm$ 0.27& 0.93\\
KP &  & 0.28 $\pm$ 0.41 & 1.00 & 0.99 $\pm$ 0.02 & 1.00 \\
BicNeuron &  & 0.39 $\pm$ 0.35 & 1.00 & 1.00 $\pm$ 0.00 & 1.00 \\
\hline
SP & \multirow{3}{*}{ringnorm} & 0.77 $\pm$ 0.30 & 1.00 & 0.32 $\pm$ 0.41& 0.99\\
KP &  & 0.00 $\pm$ 0.00 & 0.00 & 1.00 $\pm$ 0.00 & 1.00 \\
BicNeuron &  & 0.09 $\pm$ 0.05 & 0.15 & 1.00 $\pm$ 0.00 & 1.00 \\
\hline
SP & \multirow{3}{*}{singh} & 0.30 $\pm$ 0.25 & 0.80 & 1.00 $\pm$ 0.00& 1.00\\
KP &  & 0.26 $\pm$ 0.23 & 0.60 & 1.00 $\pm$ 0.00 & 1.00 \\
BicNeuron &  & 0.92 $\pm$ 0.10 & 1.00 & 0.78 $\pm$ 0.20 & 1.00 \\
\hline
SP & \multirow{3}{*}{sonar} & 0.15 $\pm$ 0.25 & 0.60 & 0.95 $\pm$ 0.14& 1.00\\
KP &  & 0.41 $\pm$ 0.39 & 1.00 & 0.95 $\pm$ 0.14 & 1.00 \\
BicNeuron &  & 0.76 $\pm$ 0.19 & 1.00 & 0.86 $\pm$ 0.17 & 1.00 \\
\hline
SP & \multirow{3}{*}{spambase} & 0.00 $\pm$ 0.00 & 0.01 & 1.00 $\pm$ 0.00& 1.00\\
KP &  & 0.04 $\pm$ 0.03 & 0.08 & 0.99 $\pm$ 0.01 & 1.00 \\
BicNeuron &  & 0.53 $\pm$ 0.35 & 0.98 & 0.79 $\pm$ 0.21 & 1.00 \\
\hline
SP & \multirow{3}{*}{tipspam} & 1.00 $\pm$ 0.00 & 1.00 & 0.02 $\pm$ 0.03& 0.09\\
KP &  & 0.00 $\pm$ 0.00 & 0.00 & 1.00 $\pm$ 0.01 & 1.00 \\
BicNeuron &  & 0.85 $\pm$ 0.07 & 0.91 & 0.54 $\pm$ 0.13 & 0.72 \\
\hline
SP & \multirow{3}{*}{twonorm} & 0.60 $\pm$ 0.18 & 0.84 & 1.00 $\pm$ 0.00& 1.00\\
KP &  & 0.00 $\pm$ 0.00 & 0.00 & 1.00 $\pm$ 0.00 & 1.00 \\
BicNeuron &  & 0.31 $\pm$ 0.09 & 0.52 & 0.91 $\pm$ 0.10 & 0.99 \\
\hline
SP & \multirow{3}{*}{wdbc} & 0.81 $\pm$ 0.10 & 0.95 & 1.00 $\pm$ 0.00& 1.00\\
KP &  & 0.14 $\pm$ 0.14 & 0.41 & 1.00 $\pm$ 0.00 & 1.00 \\
BicNeuron &  & 0.91 $\pm$ 0.07 & 1.00 & 0.96 $\pm$ 0.03 & 1.00 \\
\hline
\end{tabular}
\label{tab:accmij}
\end{table*}

\begin{table*}
\small
\centering
\setlength{\tabcolsep}{2pt}	
\caption{Average ACC and AUC results achieved by each approach for the discrimination of each pair of classes of the {\em eeg} data set. $KP_L$ and $KP_{RBF}$ denote KP with linear and RBF kernels, respectively, whereas $BN$, $BN_L$ and $BN_{RBF}$ stand for BicNeuron with standard perceptrons, with linear kernel perceptrons, and with non-linear kernel perceptrons, respectively.
The $p$-values of the Wilcoxon test (written in parentheses and rounded to two decimals) were calculated pairwise in reference to the best approach (underlined for each case) in order to assess whether their difference in performance was statistically significant ($p$-value $\leq 0.01$).}
\begin{tabular}{ccccccc}
\hline
&\multicolumn{5}{c}{ACC/AUC}\\
\hline
& A$\times$B & A$\times$C & A$\times$D & A$\times$E & B$\times$C \\
\hline
$SP$ & 0.50 $\pm$ 0.00 (0.00) & 0.72 $\pm$ 0.16 (0.00) & 0.54 $\pm$ 0.05 (0.00) & 0.56 $\pm$ 0.03 (0.00) & 0.75 $\pm$ 0.15 (0.00)\\
$KP_L$& 0.50 $\pm$ 0.00 (0.00) & 0.81 $\pm$ 0.11 (0.02) & 0.50 $\pm$ 0.00 (0.00) & 0.94 $\pm$ 0.06 (0.02) & 0.74 $\pm$ 0.17 (0.00)\\
$KP_{RBF}$& 0.72 $\pm$ 0.10 (0.08) & 0.90 $\pm$ 0.05 (0.57) & 0.83 $\pm$ 0.10 (0.19) & 0.96 $\pm$ 0.03 (0.02) & 0.93 $\pm$ 0.06 (0.00)\\
$BN$ & 0.79 $\pm$ 0.09 (0.79) & 0.88 $\pm$ 0.07 (0.13)& 0.82 $\pm$ 0.08 (0.08) & \underline{0.99 $\pm$ 0.02} (1.00) & 0.91 $\pm$ 0.07 (0.00)\\
$BN_L$& 0.78 $\pm$ 0.08 (0.54) & 0.87 $\pm$ 0.08 (0.15) & 0.79 $\pm$ 0.10 (0.05) & \underline{0.99 $\pm$ 0.02} (1.00) & \underline{1.00 $\pm$ 0.00} (1.00)\\
$BN_{RBF}$& \underline{0.81 $\pm$ 0.11} (1.00) & \underline{0.92 $\pm$ 0.04} (1.00) & \underline{0.89 $\pm$ 0.10} (1.00) & \underline{0.99 $\pm$ 0.02} (1.00) & 0.97 $\pm$ 0.05 (0.26)\\
\hline
\\
\hline
& B$\times$D & B$\times$E & C$\times$D & C$\times$E & D$\times$E \\
\hline
$SP$ & 0.51 $\pm$ 0.02 (0.00) & 0.65 $\pm$ 0.13 (0.00) & 0.49 $\pm$ 0.04 (0.00) & 0.80 $\pm$ 0.10 (0.00) & 0.72 $\pm$ 0.08 (0.00)\\
$KP_L$& 0.53 $\pm$ 0.04 (0.00) & 0.59 $\pm$ 0.12 (0.02) & 0.49 $\pm$ 0.02 (0.00) & 0.74 $\pm$ 0.10 (0.00) & 0.67 $\pm$ 0.06 (0.00)\\
$KP_{RBF}$& 0.87 $\pm$ 0.07 (0.02) & 0.90 $\pm$ 0.04 (0.06) & 0.61 $\pm$ 0.12 (0.09) & \underline{0.93 $\pm$ 0.05} (1.00) & 0.88 $\pm$ 0.07 (0.31)\\
$BN$ & 0.80 $\pm$ 0.07 (0.00) & 0.88 $\pm$ 0.09 (0.15)& 0.61 $\pm$ 0.08 (0.03) & 0.93 $\pm$ 0.04 (0.62) & 0.87 $\pm$ 0.04 (0.03)\\
$BN_L$& 0.91 $\pm$ 0.08 (0.34) & 0.93 $\pm$ 0.06 (0.62) & 0.60 $\pm$ 0.08 (0.02) & 0.92 $\pm$ 0.06 (0.62) & 0.87 $\pm$ 0.05 (0.06)\\
$BN_{RBF}$& \underline{0.95 $\pm$ 0.05} (1.00) & \underline{0.94 $\pm$ 0.02} (1.00) & \underline{0.70 $\pm$ 0.08} (1.00) & \underline{0.93 $\pm$ 0.06} (0.88) & \underline{0.92 $\pm$ 0.05} (1.00)\\
\hline
\end{tabular}
\label{tab:eeg}
\end{table*}

\subsection{Results and discussion}
\label{sec:res}

Tables~\ref{tab:acc}, \ref{tab:auc}, and \ref{tab:accmij} bring the results obtained by BicNeuron as well as SP and KP on the first 13 data sets of Table~\ref{tab:datasets}. While the first two tables provide detailed results about ACC and AUC, the former focuses on ACCm and ACCM. The main goal of this first set of experiments was to assess the quality of the discriminative subspaces discovered by BicNeuron. Since the employment of a non-linear transformation (via kernel) over these subspaces could make this assessment difficult or unclear, we refrained from adopting the RBF kernel for KP and BicNeuron by this time. Besides, for the sake of conciseness, we only show the results delivered by BicNeuron when configured with standard perceptrons, although it is important to stress that in some cases we have noticed that BicNeuron's performance could be further improved if kernel perceptrons (either with linear or non-linear kernels) were used as associated classifiers.

Before discussing qualitatively the results, it is worth explaining the notation used in Tables~\ref{tab:acc} and \ref{tab:auc} regarding the application of the Wilcoxon test for comparing the classifiers' performance (for this purpose, we adopted 0.01 as the statistical significance level). Since we are comparing the results of SP and KP against the results delivered by BicNeuron, we have decided to highlight the results of the former (SP and KP) and not the latter (BicNeuron). By this means, the notation will not become overloaded when BicNeuron is at the same time significantly better than one of the methods (cases marked in bold) and significantly worse than the other (cases marked in italics). Consider, for instance, the data set {\em twonorm}. Here, the average performance of BicNeuron ($0.61 \pm 0.01$) is significantly better than that of KP ($0.50 \pm 0.00$), so we marked the latter value in bold. On the other hand, the average performance of BicNeuron is significantly worse than that of SP ($0.80 \pm 0.09$), so we marked the latter value in {\it italics}. Notice that the relative performance of SP against KP (and vice-versa) could be deduced from their relation to BicNeuron.
 
From Tables~\ref{tab:acc} and \ref{tab:auc}, one can first notice that the three contestant approaches show some variability in performance across the different problems, with only one data set (viz. {\em chowdary}) being deemed as `easy' (almost 100\% as ACC) for all of them. Such variability is mainly due to the fact that simple linear classifiers are usually very sensitive to the underlying characteristics (such as different levels of non-linearities and noise) of the problem in sight. However, the variability of the three approaches is not the same, with their models usually showing distinct generalization capabilities, as evidenced by their fold-by-fold behavior. 

Also noteworthy is that in nine (69\%) data sets, BicNeuron delivered better average ACC and AUC results, plus one draw in one data set ({\em ringnorm}). Besides, in six cases, BicNeuron achieved the maximum possible value for ACC/AUC in at least one fold, while in 38\% and 69\% of the cases, the Wilcoxon test indicates that BicNeuron has significantly prevailed over SP and KP, respectively. These results testify the usefulness of the proposed approach, and we guess that by resorting to other biclustering algorithms (in place of the one by Huang et al.~\cite{huang2011}), better performance could be achieved in some cases.  

On the other hand, in a single problem (namely, {\em twonorm}), BicNeuron was significantly outperformed by SP, even though in this case the novel approach was still much better than KP. In this regard, it should be reminded that {\em twonorm}, as well as {\em ringnorm}, were artificially conceived by Breiman~\cite{breiman1998} while studying the bias/variance properties of single versus aggregate classifiers. While in {\em twonorm}, the optimal separating surface is an oblique plane, in {\em ringnorm}, the separating surface is a sphere. Both problems are considered as hard to approximate by simple classifier models~\cite{breiman1998}. Besides, amongst the 13 problems shown in Tables~\ref{tab:acc} and \ref{tab:auc}, {\em twonorm} and {\em ringnorm} are the ones with the lowest ratio of number of features to the number of instances, an issue that may have an impact on the quality of the biclusters induced by the biclustering algorithm.

The role of Table~\ref{tab:accmij} is to reveal the emphasis the three contestant approaches have put on the different classes and to show how the class distribution imbalance problem may affect their performance. Such results, thus, should be regarded as auxiliary to those reported in Tables~\ref{tab:acc} and \ref{tab:auc}. One can firstly notice that the approaches really vary in their emphasis on the classes, which suggests that combining their models (neurons) into more advanced architectures (such as single hidden layer feedforward networks) could be useful. Overall, BicNeuron has shown a more balanced treatment (with data set {\em tipspam} being a notorious example), which may explain why it has yielded better average ACC and AUC results in most cases. On the other hand, KP has shown high sensitiveness to high levels of class imbalance (e.g., refer to data sets \#3-\#5 and \#13), although its performance was also bad in some well-balanced problems (such as {\em singh} and {\em twonorm}).

Finally, in what concerns our second set of computational experiments, Table~\ref{tab:eeg} brings the average ACC and AUC results delivered for the 10 pairs of classes of the {\em eeg} data set. By this time, since we also aimed at assessing the effects of using a non-linear kernel over the linear subspaces discovered by BicNeuron, we have included in the contest a KP model configured with RBF kernel and BicNeuron variants configured with linear and non-linear kernel perceptrons. In total, six models were considered in the analysis. In order to differentiate the KP models, we have used the notation $KP_L$ for denoting KP with linear kernel and $KP_{RBF}$ for representing KP with RBF kernel. A similar notation was used for BicNeuron, namely, $BN_L$ and $BN_{RBF}$.

One interesting aspect to observe in these experiments is that the ACC and AUC results were the same for all models in all cases. Comparing solely the linear classifiers, the prevalence of the BicNeuron models is readily noticeable, since they have always produced better values than SP and KP with linear kernel. On the other hand, considering the non-linear models as well, one can notice that in most of the cases the BicNeuron models enhanced with RBF kernel could further improve the average performance delivered by KP configured with the same kernel. A remarkable performance was achieved while discriminating between the first and fifth classes (99\% accuracy rate) and the second and third classes (100\% accuracy rate). Overall, these results provide further evidence that the use of contrastive biclusters can indeed leverage up the predictive performance of simple models such as perceptrons while coping with non-trivial biosignal classification problems like this one~\cite{lima2011,nunes2014}.

\section{Final Remarks}
\label{sec:conclusion}

In this paper, we explored the strategy of using coherent biclusters as a means to improve the levels of accuracy and generalization exhibited by simple linear classifiers such as perceptrons. To systematically investigate the potentials of such a strategy, a novel supervised biclustering approach (BicNeuron) based on the notion of contrastive biclusters was formally devised and empirically assessed on a range of classification problems. Overall, the empirical results achieved so far show evidence about the usefulness of the linear subspaces discovered by BicNeuron for better discriminating between the classes.

A possible extension to the present work is to adapt BicNeuron to handle multiclass and multilabel data sets~\cite{prati2013} in a more straightforward manner. Moreover, we shall analyze how tolerant are the BicNeuron classifiers to noisy data~\cite{defranca2015}. 

As ongoing work, we are currently investigating the potentials of combining several BicNeuron models into the same learning framework. The idea is to better approximate the non-linear class boundaries by aggregating different local discriminative subspaces of the original data. In this context, different ways of selecting the contrastive biclusters and merging the outputs of the trained perceptrons are under consideration. Combinations of BicNeuron models with standard and kernel perceptrons should also be probed due to their complementary profiles. Finally, the whole approach could be also extended into ensemble settings by including more complex classifiers (such as other types of kernel machines~\cite{scholkopf2002}) induced over the subspaces captured by the contrastive biclusters. 

\begin{acknowledgements}
CNPq/Brazil has financially supported this work via Grants \#203295/2014-5 and \#303872/2015-2. 
\end{acknowledgements}

\bibliographystyle{spmpsci}      

\bibliography{NeuralBic}   

\begin{thebibliography}{10}
\providecommand{\url}[1]{{#1}}
\providecommand{\urlprefix}{URL }
\expandafter\ifx\csname urlstyle\endcsname\relax
  \providecommand{\doi}[1]{DOI~\discretionary{}{}{}#1}\else
  \providecommand{\doi}{DOI~\discretionary{}{}{}\begingroup
  \urlstyle{rm}\Url}\fi

\bibitem{alpaydin2014}
Alpaydin, E.: Introduction to Machine Learning, 3rd edn.
\newblock MIT Press (2014)

\bibitem{breiman1998}
Breiman, L.: Arcing classifiers.
\newblock Ann. Statist. \textbf{26}(3), 801--824 (1998)

\bibitem{cheng2000}
Cheng, Y., Church, G.M.: Biclustering of expression data.
\newblock In: Proc. of the 8th Int. Conf. on Intelligent Systems for Molecular
  Biology, pp. 93--103 (2000)

\bibitem{costa2013}
Costa, H., Benevenuto, F., Merschmann, L.H.C.: Detecting tip spam in
  location-based social networks.
\newblock In: Proc. of the 28th ACM Symp. on Applied Computing (SAC) (2013).
\newblock
  \urlprefix\url{http://homepages.dcc.ufmg.br/~fabricio/spamcollectionLBSN.html}

\bibitem{delve1998}
{DELVE}: Delve repository (1998).
\newblock \urlprefix\url{http://www.cs.toronto.edu/~delve/}

\bibitem{demsar2006}
Dem\v{s}ar, J.: Statistical comparisons of classifiers over multiple data sets.
\newblock J. Mach. Learn. Res. \textbf{7}, 1--30 (2006)

\bibitem{dhillon2001}
Dhillon, I.S.: Co-clustering documents and words using bipartite spectral graph
  partitioning.
\newblock In: Proc. of the 7th Int. Conf. on Knowledge Discovery and Data
  Mining, pp. 269--274 (2001)

\bibitem{dhillon2003}
Dhillon, I.S., Mallela, D.S., Modha, S.: Information theoretic co-clustering.
\newblock In: Proc. of the ACM SIGKDD Conference on Knowledge Discovery and
  Data Mining, pp. 89--98. Springer (2003)

\bibitem{dong2012}
Dong, G., Bailey, J. (eds.): Contrast Data Mining: Concepts, Algorithms, and
  Applications.
\newblock Chapman \& Hall/CRC Press (2012)

\bibitem{duch2012}
Duch, W., Jankowski, N., Maszczyk, T.: Make it cheap: {L}earning with ${O}(nd)$
  complexity.
\newblock In: Proc. of 2012 IEEE International Joint Conference on Neural
  Networks (IJCNN), pp. 1--4 (2012)

\bibitem{fang2010}
Fang, G., Kuang, R., Pandey, G., Steinbach, M., Myers, C.L., Kumar, V.:
  Subspace differential coexpression analysis: {P}roblem definition and a
  general approach.
\newblock In: Proc. of 15th Pacific Symp. on Biocomp., pp. 145--156 (2010)

\bibitem{defranca2015}
de~Fran{\c{c}}a, F.O., Coelho, A.L.V.: A biclustering approach for
  classification with mislabeled data.
\newblock Expert Syst. Appl. \textbf{42}(12), 5065--5075 (2015)

\bibitem{defranca2013}
de~Fran{\c{c}}a, F.O., Coelho, G.P., Von~Zuben, F.J.: Predicting missing values
  with biclustering: A coherence-based approach.
\newblock Pattern Recogn. \textbf{46}(5), 1255--1266 (2013)

\bibitem{freitas2013}
Freitas, A., Ayadi, W., Elloumi, M., Oliveira, J.L., Hao, J.K.: Survey on
  biclustering of gene expression data.
\newblock In: Biological Knowledge Discovery Handbook: Preprocessing, Mining,
  and Postprocessing of Biological Data. Wiley (2013)

\bibitem{freund1999}
Freund, Y., Schapire, R.E.: Large margin classification using the perceptron
  algorithm.
\newblock Machine Learn. \textbf{37}(3), 277--296 (1999)

\bibitem{han2009}
Han, L., Yan, H.: A fuzzy biclustering algorithm for social annotations.
\newblock Journal of Information Science \textbf{35}, 426--438 (2009)

\bibitem{huang2011}
Huang, Q., Tao, D., Li, X., Jin, L., Wei, G.: Exploiting local coherent
  patterns for unsupervised feature ranking.
\newblock IEEE Trans. Syst., Man, Cybern., Part B: Cybernetics \textbf{41}(6),
  1471--1482 (2011)

\bibitem{huang2015}
Huang, Q., Wang, T., Tao, D., Li, X.: Biclustering learning of trading rules.
\newblock IEEE Transactions on Cybernetics \textbf{45}(10), 2287--2298 (2015)

\bibitem{jain2010}
Jain, A.K.: Data clustering: 50 years beyond $k$-means.
\newblock Pattern Recogn. Lett. \textbf{31}, 651--666 (2010)

\bibitem{lichman2013}
Lichman, M.: {UCI} machine learning repository (2013).
\newblock \urlprefix\url{http://archive.ics.uci.edu/ml}

\bibitem{lima2011}
Lima, C.A.M., Coelho, A.L.V.: Kernel machines for epilepsy diagnosis via {EEG}
  signal classification: A comparative study.
\newblock Artif. Intell. Med. \textbf{53}, 83--95 (2011)

\bibitem{madeira2004}
Madeira, S., Oliveira, A.L.: Biclustering algorithms for biological data
  analysis: {A} survey.
\newblock IEEE/ACM Trans. Comput. Biol. Bioinf. \textbf{1}, 24--45 (2004)

\bibitem{nunes2014}
Nunes, T.M., Coelho, A.L.V., Lima, C.A.M., Papa, J.P., de~Albuquerque, V.H.C.:
  {EEG} signal classification for epilepsy diagnosis via optimum path forest --
  {A} systematic assessment.
\newblock Neurocomputing \textbf{136}, 103--123 (2014)

\bibitem{odibat2014}
Odibat, O., Reddy, C.K.: Efficient mining of discriminative co-clusters from
  gene expression data.
\newblock Knowl. Inf. Syst. \textbf{41}, 667--696 (2014)

\bibitem{odibat2010}
Odibat, O., Reddy, C.K., Giroux, C.N.: Differential biclustering for gene
  expression analysis.
\newblock In: Proc. of the ACM Conference on Bioinformatics and Computational
  Biology, pp. 275--284 (2010)

\bibitem{prati2013}
Prati, R.C., de~Fran{\c c}a, F.O.: Extending features for multilabel
  classification with swarm biclustering.
\newblock In: Proc. of 2013 IEEE Congress on Evolutionary Computation (CEC),
  pp. 2964--2971. IEEE (2013)

\bibitem{scholkopf2002}
Sch{\"o}lkopf, B., Smola, A.: Learning with Kernels.
\newblock MIT Press (2002)

\bibitem{desouto2008}
de~Souto, M.C.P., Costa, I.G., de~Araujo, D.S.A., Ludermir, T.B., Schliep, A.:
  Clustering cancer gene expression data: {A} comparative study.
\newblock BMC Bioinformatics \textbf{9}, 497 (2008)

\bibitem{subasi2008}
Subasi, A.: {EEG} signal classification using wavelet feature extraction and a
  mixture of expert model.
\newblock Expert Syst. Appl. \textbf{32}, 1084---1093 (2007)

\bibitem{tang2009}
Tang, Y.Y.: Wavelet Theory Approach to Pattern Recognition.
\newblock World Sci. Publishing (2009)

\bibitem{wang2013}
Wang, M., Shang, X., Li, X., Liu, W., Li, Z.: Efficient mining differential
  co-expression biclusters in microarray datasets.
\newblock Gene \textbf{518}, 59--69 (2013)

\bibitem{zhao2012}
Zhao, H., Liew, A.W.C., Wang, D.Z., Yan, H.: Biclustering analysis for pattern
  discovery: {C}urrent techniques, comparative studies and applications.
\newblock Current Bioinformatics \textbf{7}(43--55) (2012)

\end{thebibliography}

\end{document}